\documentclass{article}
\usepackage{PRIMEarxiv}

\usepackage[utf8]{inputenc} 
\usepackage{hyperref}       
\usepackage{url}            
\usepackage{booktabs}       
\usepackage{amsfonts}       
\usepackage{nicefrac}       
\usepackage{microtype}      
\usepackage{lipsum}
\usepackage{fancyhdr}       
\usepackage{graphicx}       
\graphicspath{{img/}}     
\usepackage{amsmath}
\usepackage{microtype}      
\usepackage{cleveref}       
\usepackage{lipsum}         
\usepackage{graphicx}
\usepackage[sort&compress,numbers]{natbib}
\usepackage{doi}
\usepackage{enumerate}
\usepackage{amssymb,enumitem}
\usepackage{multirow}
\usepackage{xcolor}
\usepackage{longtable}
\usepackage{algorithm}
\usepackage{algpseudocode}
\usepackage{array}
\makeatletter
\def\algbackskip{\hskip-\ALG@thistlm}

\usepackage{subcaption}
\usepackage{listings}

\title{Advancing Wound Filling Extraction on 3D Faces: Auto-Segmentation and Wound Face Regeneration Approach}
\author{
Duong Q. Nguyen$^{1}$, Thinh D. Le$^{2}$, Phuong D. Nguyen$^{2}$, Nga T.K. Le$^{3}$, H. Nguyen-Xuan$^{2,}$\thanks{Corresponding author: CIRTECH Institute, HUTECH University (Email: ngx.hung@hutech.edu.vn).}  \\
$^{1}$Department of Mathematics and Statistics, Quy Nhon University, Quy Nhon City, Viet Nam\\
$^{2}$CIRTECH Institute, HUTECH University, Ho Chi Minh City, Viet Nam \\
$^{3}$Applied Research Institute for Science and Technology, Quy Nhon University, Vietnam
}
\begin{document}
\maketitle
\begin{abstract}
Facial wound segmentation plays a crucial role in preoperative planning and optimizing patient outcomes in various medical applications. In this paper, we propose an efficient approach for automating 3D facial wound segmentation using a two-stream graph convolutional network. Our method leverages the Cir3D-FaIR dataset and addresses the challenge of data imbalance through extensive experimentation with different loss functions. To achieve accurate segmentation, we conducted thorough experiments and selected a high-performing model from the trained models. The selected model demonstrates exceptional segmentation performance for complex 3D facial wounds. Furthermore, based on the segmentation model, we propose an improved approach for extracting 3D facial wound fillers and compare it to the results of the previous study. Our method achieved a remarkable accuracy of 0.9999986\% on the test suite, surpassing the performance of the previous method. From this result, we use 3D printing technology to illustrate the shape of the wound filling. The outcomes of this study have significant implications for physicians involved in preoperative planning and intervention design. By automating facial wound segmentation and improving the accuracy of wound-filling extraction, our approach can assist in carefully assessing and optimizing interventions, leading to enhanced patient outcomes. Additionally, it contributes to advancing facial reconstruction techniques by utilizing machine learning and 3D bioprinting for printing skin tissue implants. Our source code is available at \url{https://github.com/SIMOGroup/WoundFilling3D}.
\end{abstract}


\section{Introduction}
Nowadays, people are injured by traffic accidents, occupational accidents, birth defects, diseases that have made them lose a part of their body. In which, defects when injured in the head and face areas account for a relatively high rate \cite{Susanne_2023}. Wound regeneration is an important aspect of medical care, aimed at restoring damaged tissues and promoting wound healing in patients with complex wounds \cite{han2023innovations}. However, the treatment of craniofacial and facial defects can be challenging due to the many specific requirements of the tissue and the complexity of the anatomical structure of that region \cite{Nyberg_2016}. Traditional methods used for wound reconstruction often involve grafting techniques using automated grafts (from the patient's own body) or allogeneic grafts (from a donor) \cite {Alexander_2021}. However, these methods have limitations such as availability, donor morbidity, and potential for rejection. In recent years, the development of additive manufacturing technology has promoted the creation of advanced techniques in several healthcare industries \cite{Atabak_2022, Greymi_2022, Relano_2021}. The implementation of 3D printing technology in the preoperative phase enables clinicians to establish a meticulous surgical strategy by generating an anatomical model that accurately reflects the patient's unique anatomy. This approach facilitates the development of customized drilling and cutting instructions, precisely tailored to the patient's specific anatomical features, thereby accommodating the potential incorporation of a pre-formed implant \cite{Hitesh_2018}. Moreover, the integration of 3D printing technology and biomaterials assumes a pivotal role in advancing innovative remedies within the field of regenerative medicine, addressing the pressing demand for novel therapeutic modalities \cite{Tack_2016, Don_16, Nanbo_2021, Wallace_2023}. The significance of wound reconstruction using 3D bioprinting in the domain of regenerative medicine is underscored by several key highlights, as outlined below:
\begin{itemize}
\item[-] \textbf{Customization and Precision}: 3D bioprinting allows for the creation of patient-specific constructs, tailored to match the individual's wound geometry and requirements. This level of customization ensures a better fit and promotes improved healing outcomes.
\item[-] \textbf{Tissue Regeneration}: The ability to fabricate living tissues using 3D bioprinting holds great promise for wound reconstruction. The technique enables the deposition of cells and growth factors in a controlled manner, facilitating tissue regeneration and functional restoration \cite{Cubo_2017, Chuang_2021}.
\item[-] \textbf{Reduced Donor Dependency}: The scarcity of donor tissues and the associated risks of graft rejection are significant challenges in traditional wound reconstruction methods. 3D bioprinting can alleviate these limitations by providing an alternative approach that relies on the patient's own cells or bioinks derived from natural or synthetic sources \cite{Jain_2022}.
\item[-] \textbf{Complex Wound Healing}: Certain wounds, such as large burns, chronic ulcers, or extensive tissue loss, pose significant challenges to conventional wound reconstruction methods. 3D bioprinting offers the potential to address these complex wound scenarios by creating intricate tissue architectures that closely resemble native tissues.
\item[-] \textbf{Accelerated Healing}: By precisely designing the structural and cellular components of the printed constructs, 3D bioprinting can potentially enhance the healing process. This technology can incorporate growth factors, bioactive molecules, and other therapeutic agents, creating an environment that stimulates tissue regeneration and accelerates wound healing \cite{Li_2021}.
\end{itemize}
Consequently, 3D bioprinting technology presents a promising avenue for enhancing craniofacial reconstruction modalities in individuals afflicted by head trauma.

Wound dimensions, including length, width, and depth, are crucial parameters for assessing wound healing progress and guiding appropriate treatment interventions \cite{Grey_2006}. For effective facial reconstruction, measuring the dimensions of a wound accurately can pose significant challenges in clinical and scientific settings \cite{Musa_2022}. Firstly, wound irregularity presents a common obstacle. Wounds rarely exhibit regular shapes, often characterized by uneven edges, irregular contours, or irregular surfaces. Such irregularity complicates defining clear boundaries and determining consistent reference points for measurement. Secondly, wound depth measurement proves challenging due to undermined tissue or tunnels. These features, commonly found in chronic or complex wounds, can extend beneath the surface, making it difficult to assess the wound's true depth accurately. Furthermore, the presence of necrotic tissue or excessive exudate can obscure the wound bed, further hindering depth measurement. Additionally, wound moisture and fluid dynamics pose significant difficulties. Wound exudate, which may vary in viscosity and volume, can accumulate and distort measurements. Excessive moisture or the presence of dressing materials can alter the wound's appearance, potentially leading to inaccurate measurements. Moreover, the lack of standardization in wound measurement techniques and tools adds to the complexity.

\begin{figure}[!h]
\centering
\includegraphics[scale=0.3]{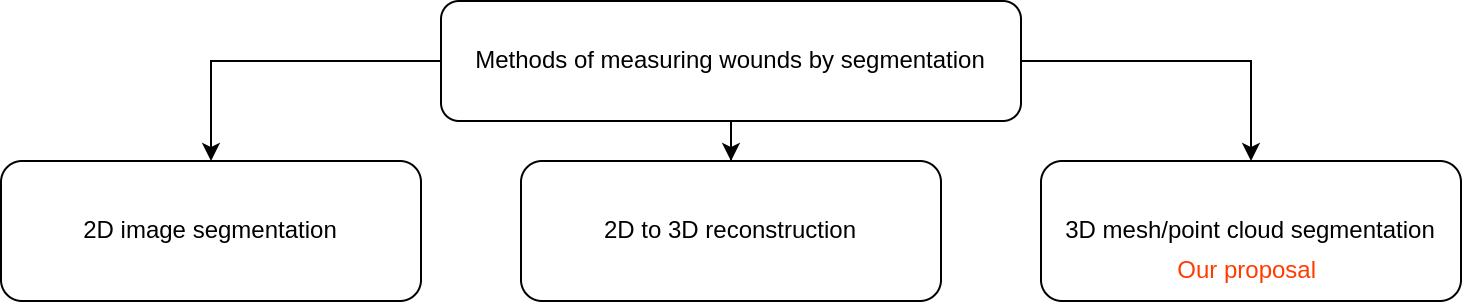}
\caption{Methods of using deep learning in wound measurement by segmentation}
\label{method_segmentation}
\end{figure}

Currently, deep learning has rapidly advanced in computer vision and medical imaging, emerging as the predominant technique for wound image segmentation. \cite{Geert_2017, Gaetano_2022, Juhong_2021}. Based on the characteristics of the input data \cite{Zhang_2022, Anisuzzaman_2022}, three deep learning methods are used for segmentation and wound measurement, as shown in Fig. \ref{method_segmentation}. The study of Anisuzzaman et al. \cite{Anisuzzaman_2022} presents case studies of these three methods. The methods used to segment the wound based on the characteristics of the input data are as follows
\begin{itemize}
\item[-] \textbf{2D image segmentation}: Deep learning methods in 2D for wound segmentation offer several advantages. Firstly, they are a well-established and widely used technique in the field. Additionally, large annotated 2D wound segmentation datasets are available, facilitating model training and evaluation. These methods exhibit efficient computational processing compared to their 3D counterparts, enabling faster inference times and improved scalability. Furthermore, deep learning architectures, such as convolutional neural networks, can be leveraged for effective feature extraction, enhancing the accuracy of segmentation results. However, certain disadvantages are associated with deep learning methods in 2D for wound segmentation. One limitation is the lack of depth information, which can restrict segmentation accuracy, particularly for complex wounds with intricate shapes and depth variations. Additionally, capturing the wound's full spatial context and shape information can be challenging in 2D, as depth cues are not explicitly available. Furthermore, these methods are susceptible to variations in lighting conditions, image quality, and perspectives, which can introduce noise and affect the segmentation performance. 
\item[-] \textbf{2D to 3D reconstruction}: By incorporating depth information, the conversion to 3D enables a better capture of wounds' shape and spatial characteristics, facilitating a more comprehensive analysis. Moreover, there is a potential for improved segmentation accuracy compared to 2D methods, as the additional dimension can provide richer information for delineating complex wound boundaries. Nevertheless, certain disadvantages are associated with converting from 2D to 3D for wound segmentation. The conversion process itself may introduce artifacts and distortions in the resulting 3D representation, which can impact the accuracy of the segmentation. Additionally, this approach necessitates additional computational resources and time due to the complexity of converting 2D data into a 3D representation \cite{Xiaorui_2023}. Furthermore, the converted 3D method may not completely overcome the limitations of the 2D method.
\item[-]\textbf{3D mesh or point cloud segmentation}: Directly extracting wound segmentation from 3D data (mesh/point cloud) offers several advantages. One notable advantage is the retention of complete 3D information on the wound, enabling accurate and precise segmentation. By working directly with the 3D data, this method effectively captures the wound's intricate shape, volume, and depth details, surpassing the capabilities of both 2D approaches and converted 3D methods. Furthermore, the direct utilization of 3D data allows for a comprehensive analysis of the wound's spatial characteristics, facilitating a deeper understanding of its structure and morphology. 
\end{itemize}
Hence, employing a 3D (mesh or point cloud) segmentation method on specialized 3D data, such as those obtained from 3D scanners or depth sensors, can significantly improve accuracy compared to the other two methods. The use of specialized 3D imaging technologies enables the capture of shape, volume, and depth details with higher fidelity and accuracy \cite{Shah_2015}. Consequently, the segmentation results obtained from this method are expected to provide a more precise delineation of wound boundaries and a more accurate assessment of wound characteristics. Therefore, this method can enhance wound segmentation accuracy and advance wound assessment techniques.

Besides, facial wounds and defects present unique challenges in reconstructive surgery, requiring accurate localization of the wound and precise estimation of the defect area \cite{Anish_2023}. The advent of 3D imaging technologies has revolutionized the field, enabling detailed capture of facial structures. However, reconstructing a complete face from a 3D model with a wound remains a complex task that demands advanced computational methods. Accurately reconstructing facial defects is crucial for surgical planning, as it provides essential information for appropriate interventions and enhances patient outcomes \cite{Michael_2019}. Some prominent studies, such as Sutradhar et al. \cite{Sutradhar_2016} utilized a unique approach based on topology optimization to create patient-specific craniofacial implants using 3D printing technology; Nuseir et al. \cite{Nuseir_2018} proposed the utilization of direct 3D printing for the fabrication of a pliable nasal prosthesis, accompanied by the introduction of an optimized digital workflow spanning from the scanning process to the achievement of an appropriate fit; and some other prominent studies presented in survey studies such as \cite{Ghai_2018, Salah_2020}. However, these methods often require a lot of manual intervention and are prone to subjectivity and variability. To solve this problem, the method proposed in \cite{Phuong_2023, Phuong_2023_self} leverages the power of modeling \cite{zhou_fully_2020} to automate the process of 3D facial reconstruction with wounds, minimizing human error and improving efficiency. To extract the filling for the wound, the study \cite{Phuong_2023} proposed the method of using the reconstructed 3D face and the 3D face of the patient without the wound. This method is called \textit{outlier extraction} by the authors. These advancements can be leveraged to expedite surgical procedures, enhance precision, and augment patient outcomes, thereby propelling the progression of technology-driven studies on facial tissue reconstruction, particularly in bio 3D printing. However, this method still has some limitations as follows
\begin{itemize}
\item[-] The method of extracting filler for the wound after 3D facial reconstruction has not yet reached high accuracy.
\item[-] In order to extract the wound filling, the method proposed by [12] necessitated the availability of the patient's pre-injury 3D facial ground truth. This requirement represents a significant limitation of the proposed wound filling extraction approach, as obtaining the patient's pre-injury 3D facial data is challenging in real-world clinical settings.
\end{itemize}
To overcome these limitations, the present study aims to address the following objective:
\begin{itemize}
\item[-]  Train the 3D facial wound container segmentation automatic model using a variety of appropriate loss functions to solve the data imbalance problem.
\item[-] Propose an efficient approach to extract the 3D facial wound filling by leveraging the face regeneration model in the study \cite{Phuong_2023} combined with the wound segmentation model.
\item[-] Evaluate the experimental results of our proposed method and the method described in the study by Phuong et al. \cite{Phuong_2023}. One case study will be selected to be illustrated through 3D printing.
\end{itemize}
\section{Methodology}
\label{Recommended_method}
In recent years, 3D segmentation methods have made notable advancements in various fields, including computer vision and medical imaging \cite{Abubakar_2022}. Prominent approaches such as PointNet \cite{Charles_2017}, PointNet++ \cite{Qi_2017}, PointCNN \cite{Li_2018}, MeshSegNet \cite{Lian2019MeshSNetDM}, and DGCNN \cite{Wang_2019} have demonstrated their efficacy in segmenting 3D data. Among these methods, the two-stream graph convolutional network (TSGCNet) \cite{Zhao_2022} for 3D segmentation stands out as an exceptional technique, demonstrating outstanding performance and potential in the field. This network exploits many of the powerful geometric features available in mesh to perform segmentation tasks. Consequently, in this study, we have chosen this model as the focal point to investigate its applicability and effectiveness in the context of our research objectives. In \cite{Zhao_2022}, the proposed methodology utilizes two parallel streams, namely the $C$ stream and the $N$ stream. To extract high-level geometric representations from the coordinates and normal vectors, TSGCNet incorporate input-specific graph-learning layers. Subsequently, the features acquired from these two complementary streams are amalgamated in the feature-fusion branch to facilitate the acquisition of discriminative multi-view representations, specifically for segmentation purposes. We introduce the details of each active flow of the two-stream graph convolutional network.

\subsection{Architecture of the $C$-stream}
The $C$-stream is designed to capture the essential topological characteristics derived from the coordinates of all vertices. The $C$-stream receives an input denoted as $\mathbf{F}_\mathbf{c}^{0}$, which is an $M \times 12$ matrix representing the coordinates. Each row of this matrix represents a node, and the columns correspond to the coordinates of the cell in a three-dimensional space. This stream incorporates an input-transformer module to align the input data with a canonical space. This module comprises shared Multilayer Perceptrons (MLPs) across nodes, as previously described by Charles et al. \cite{Charles_2017}. The objective of this module is to acquire the parameters of an affine transformation matrix, represented as $\mathbf{T} \in \mathbb{R}^{12 \times 12}$. The affine transformation is applied to the original coordinate matrix $\mathbf{F}^0_\mathbf{c}$ using matrix multiplication. The resulting transformed matrix is represented as $\hat{\mathbf{F}^0_\mathbf{c}}$, and the transformation operation can be expressed as:
\begin{equation}
\hat{\mathbf{F}^0_\mathbf{c}} = \mathbf{F}^0_\mathbf{c} \mathbf{T}.
\end{equation}
By multiplying the input matrix $\mathbf{F}^0_\mathbf{c}$ with the affine transformation matrix $\mathbf{T}$, the $C$-stream aligns the coordinates of the nodes to a standard reference or canonical space. The alignment operation plays a vital role in maintaining the consistency and stability during the extraction of geometric features in the subsequent layers of the network. By stabilizing the feature extraction process, the network can more effectively capture essential topological information from the input data. The $C$-stream progressively integrates a consecutive set of graph-attention layers along the forward path to systematically exploit multi-scale geometric attributes derived from the coordinate aspect. Let $\mathbf{F}^l_c \in \mathbb{R}^{M \times d}$ represent the matrix learned by the $(l-1)$-th graph attention layer. The row vector $\mathbf{f}_i^l \in \mathbb{R}^d$ within $\mathbf{F}^l_c$ signifies the representation of the $i$-th node $p_i$. Subsequently, high-level geometric representations $\mathbf{F}_c^{l+1} \in \mathbb{R}^{M\times k}$ is further extracted by the following $l-$th graph-attention layer in four steps.

\textit{Step 1. Constructing the Dynamic KNN Graph}: A graph $G$($V$, $E$) is designed in terms of $\mathbf{F}_c^l$, where $V = \{p_1,p_2,...,p_M\}$ signifies the collection of $M$ nodes and $E$ denotes the corresponding set of edges which determined by the K-nearest neighbors (KNN) connectivity. It should be emphasized that each node $p_i \in V$ merely connects to its KNNs, which are represented by the symbol $\mathcal{N}(i)$.

\textit{Step 2. Local Information Calibration}: For each $p_i\in V$, we update the representation $\mathbf{f}_{ij}^l$ of the $j$-th nearest neighbor $p_{ij} \in \mathcal{N}(i)$  by integrating its own representation, more precisely:
	\begin{equation}
		\hat{\mathbf{f}}_{ij}^l = \mathbf{MLP}^l(\mathbf{f}_i^{l}\oplus\mathbf{f}_{ij}^{l}),
	\label{step2}
	\end{equation}
	where $\forall p_{ij} \in \mathcal{N}(i)$ and $\oplus$ represents the operation for channel-wise concatenation. This operation combines the channels from multiple arrays into a single array, preserving the information from each channel (a concatenation operation is just a stacking operation). The $p_{ij}$ can be the nearest neighbor of many central nodes $p_i$. Therefore, the information provided by $p_{ij}$ via the $\hat{\mathbf{f}}_{ij}^l$ can be more consistent with the central node $p_i.$

\textit{Step 3. Estimating Attention Weights}: The estimation of attention weights for the neighborhood of each node $p_i$ is accomplished by utilizing a lightweight network shared among nodes. This network is designed to capture the local geometric characteristics. To be more precise,  
\begin{equation}
		\gamma_{i j}^l=\sigma\left((\mathbf{f}_i^l-\mathbf{f}_{i j}^l) \oplus \mathbf{f}_{i j}^l\right), \quad \forall p_{i j} \in \mathcal{N}(i),
\end{equation} where $\gamma_{i j}^l\in\mathbb{R}^k$ is attention weight of neighbor $ p_{ij} $
in the $l$-th layer, $\sigma(\cdot)$ is implemented as a lightweight MLP.

\textit{Step 4.  Aggregating Neighborhood Information:} The feature aggregation in the $l$-th layer is mathematically expressed as:
\begin{equation}
	\mathbf{f}_i^{l+1}=\sum_{p_{i j} \in \mathcal{N}(i)} \gamma_{i j}^l \odot \hat{\mathbf{f}}_{i j}^l,
\end{equation}
where $\odot$ denotes the element-wise multiplication operation performed between two feature vectors.
\subsection{Architecture of the $N$-stream}
The $C$-stream primarily focuses on extracting the basic structure and topology from the coordinates of the nodes. While it can capture general geometric information, it lacks the sensitivity to distinguish subtle boundaries between adjacent nodes with different classes (e.g., the boundary between the injured and non-injured areas). To overcome this limitation, the $N$-stream is introduced. The $N$-stream is designed to extract boundary representations based on the normal vectors associated with the nodes. Normal vectors provide information about the orientation and surface properties of the nodes. By integrating the normal vectors of all nodes as inputs, the $N$-stream can effectively capture detailed and precise boundary information during the learning process. Before extracting the higher-level geometric features, a transformation is applied to return the normal vectors to the same normal space. This normalization helps to ensure consistency in the representation of normal vectors and facilitates further processing. The $N$-stream is constrained to utilize the same KNN graphs constructed in the $C$-stream to learn boundary representations within local regions and mitigate interference from distant nodes sharing similar normal vectors belonging to distinct classes. This constraint ensures that the $N$-stream focuses on capturing meaningful boundary information while accounting for the local context of each node. This shared KNN graph ensures that the relationships between nodes captured by both streams are consistent and aligned. By sharing the KNN graphs, the $N$-stream can focus on local regions and their associated boundaries. Using graph max-pooling layers in the $N$-stream distinguishes it from the graph-attention layers employed in the $C$-stream. This differentiation is essential as the normal vectors encompass unique geometric information that differs from the coordinates of the nodes. Because the normal vector carries only geometry information, the $N$ stream prefers to use max-pooling layers instead of graph-attention layers as in the $C$ stream. Similar to the $C$ stream, the $l$-th graph max-pooling layer modifies the information locally for each $p_i$ node by updating $\mathbf{f}_i^l$ to $\hat{\mathbf {f}}_i^l$ based on Eq. (\ref{step2}). In order to create the boundary representation for each center $p_i$, channel-wise max-pooling is applied to aggregate the $\hat{\mathbf {f}}_i^l$ corrected features given by the following formula
\begin{equation}
	\mathbf{f}_i^{l+1}=\operatorname{maxpooling}\left\{\hat{\mathbf{f}}_{i j}^l, \forall p_{i j} \in N(i)\right\}.
\end{equation}

\subsection{Combination of features}
Research \cite{Zhao_2022} used MLP to represent a high-level view by concatenating features from each layer $\mathbf{F}_c^l$ and $\mathbf{F}_n^l$, denoted as
\begin{align}
& \mathbf{F}_{\mathbf{c}}=\mathbf{M L P}_{\mathbf{c}}\left(\mathbf{F}_{\mathbf{c}}^1 \oplus \mathbf{F}_{\mathbf{c}}^2 \oplus \mathbf{F}_{\mathbf{c}}^3\right), \\
& \mathbf{F}_{\mathbf{n}}=\mathbf{M L P}_{\mathbf{n}}\left(\mathbf{F}_{\mathbf{n}}^1 \oplus \mathbf{F}_{\mathbf{n}}^2 \oplus \mathbf{F}_{\mathbf{n}}^3\right).
\end{align}
These representations help capture information from local to global features in each stream. Subsequently $\mathbf{F}_{\mathbf{c}}$ and $\mathbf{F}_{\mathbf{n}}$ are normalized to the mesh according to the following formula
\begin{align}
	& \vartheta_{\mathbf{c}}=\frac{\left|\mathbf{f}_{\mathbf{n}}^i\right|}{\left|\mathbf{f}_{\mathbf{c}}^i\right|+\left|\mathbf{f}_{\mathbf{n}}^i\right|}, \label{normali}\\
	& \vartheta_{\mathbf{n}}=\frac{\left|\mathbf{f}_{\mathbf{c}}^i\right|}{\left|\mathbf{f}_{\mathbf{c}}^i\right|+\left|\mathbf{f}_{\mathbf{n}}^i\right|},
\end{align}
where $\mathbf{f}_{\mathbf{c}}^i \in \mathbf{F}_{\mathbf{c}}$ and $\mathbf{f}_{\mathbf{n}}^i \in \mathbf{F}_{\mathbf{n}}$. After that, the feature vectors $\mathbf{f}_{\mathbf{c}}^i$ and $\mathbf{f}_{\mathbf{n}}^i$ calculated after normalization have the following form
\begin{align}
	\hat{\mathbf{f}_{\mathbf{c}}^i}=\vartheta_{\mathbf{c}} \mathbf{f}_{\mathbf{c}}^i, \\
	\hat{\mathbf{f}_{\mathbf{n}}^i}=\vartheta_{\mathbf{n}} \mathbf{f}_{\mathbf{n}}^i .
\label{setnormali}
\end{align}
Eqs. (\ref{normali})-(\ref{setnormali}) are the purpose of bridging the gap between features $\hat{\mathbf{f}_{\mathbf{c}}^i}$ and $\hat{\mathbf{f}_{\mathbf{n}}^i}$. Nevertheless, feature $\hat{\mathbf{f}_{\mathbf{n}}^i}$ uses the max-pooling operator, which can result in a higher numerical magnitude compared to feature $\hat{\mathbf{f}_{\mathbf{c}}^i}$. To solve this problem, the self-attention mechanism was applied to balance the contributions of $\hat{\mathbf{f}_{\mathbf{c}}^i}$ and $\hat{\mathbf{f}_{\mathbf{n}}^i}$. Specifically, $\mathbf{M L P}_{\mathbf{A t t}}$ is a lightweight MLP, the calculation of the self-attention weight is determined by the following equation:
\begin{equation}
\beta_i=\mathbf{M L P}_{\mathbf{A t t}}\left(\hat{\mathbf{f}_{\mathbf{c}}^i} \oplus \hat{\mathbf{f}_{\mathbf{n}}^i}\right),
\end{equation}
where $\beta_i$ has the same shape as $\left(\hat{\mathbf{f}_{\mathbf{c}}^i} \oplus \hat{\mathbf{f}_{\mathbf{n}}^i}\right)$. 
The quantification of the multi-view geometric feature representation for $p_i$ is expressed as follows:
\begin{equation}
\hat{\mathbf{f}^i}=\beta_i \odot\left(\hat{\mathbf{f}{\mathbf{c}}^i} \oplus \hat{\mathbf{f}{\mathbf{n}}^i}\right),
\label{atpt}
\end{equation}
where the symbol $\odot$ denotes the element-wise multiplication operation. Then, the feature matrix $\hat{\mathbf{F}} = (\hat{\mathbf{f}^{1}}, \hat{\mathbf{f}^2}, \ldots, \hat{\mathbf{f}^M})$ is defined represents the multi-view features for all nodes. The feature matrix undergoes a transformation through a MLP, resulting in the generation of an $M \times C$ matrix denoted as $\mathbf{P}$. Each row of $\mathbf{P}$ corresponds to the probabilities of a particular node belonging to $C$ distinct classes.

\section{Experimental tests}
This section introduces our experimental setup, including data set information and implementation details.
\subsection{Dataset}
\begin{figure}[!h]
\centering
\includegraphics[scale=0.3]{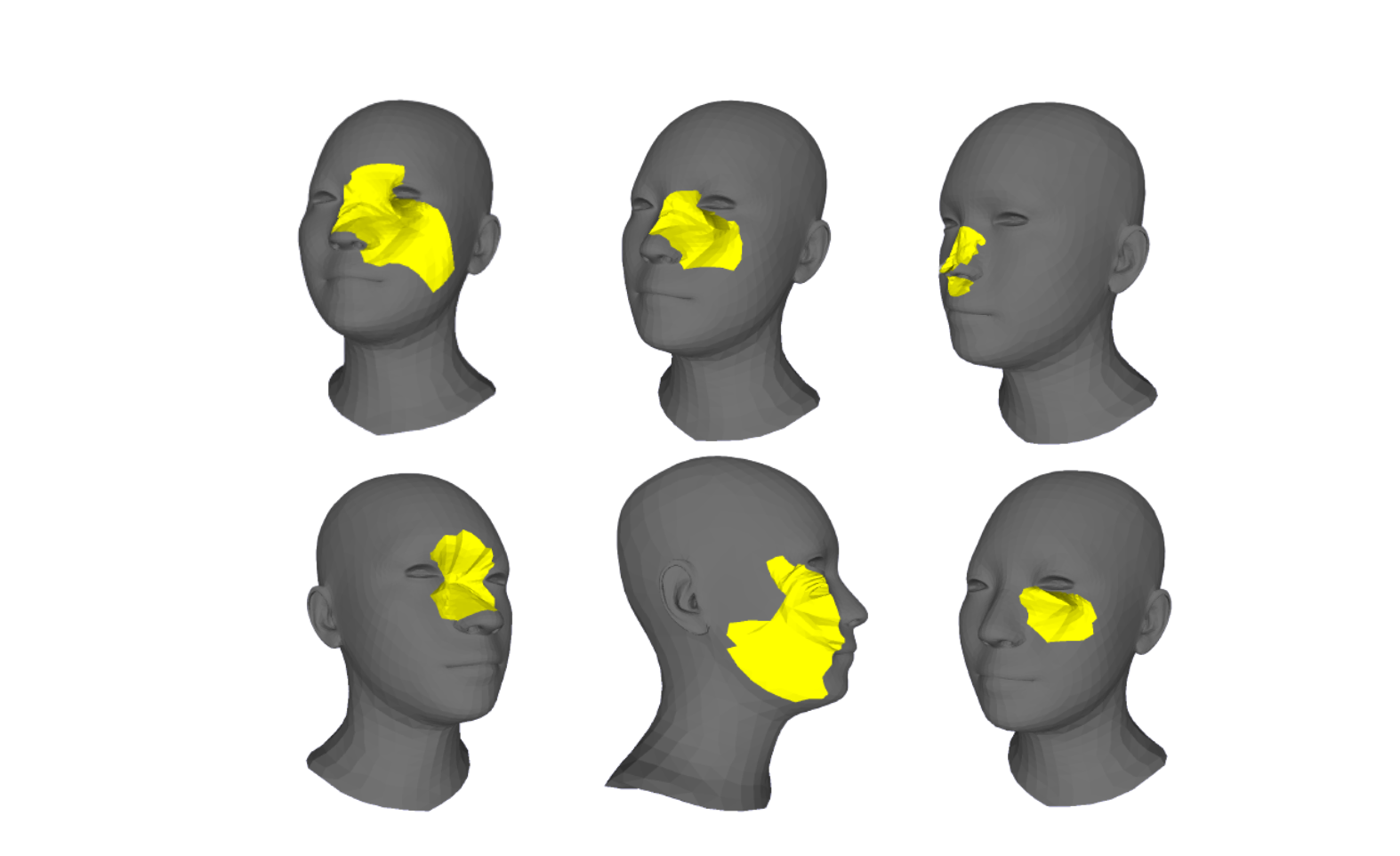}
\caption{Some illustrations of the face dataset with wounds}
\label{sample_data}
\end{figure}

We use a dataset of 3D faces of craniofacial injuries named Cir3D-FaIR \cite{Phuong_2023}. The dataset utilized in this study is generated through simulation within a virtual environment, replicating realistic facial wound locations. A set of 3,678 3D mesh representations of uninjured human faces is utilized to simulate facial wounds. Specifically, each face in the dataset is simulated with ten distinct wound locations. Consequently, the dataset has 40,458 human head meshes, encompassing uninjured faces and wounded in many different positions. Each 3D face mesh consists of 15,000 faces and is labeled according to the face location of the mesh, indicating specifically the presence of the wounds. This simulation dataset has been expert physicians to assess the complexity associated with the injuries. Figure \ref{sample_data} showcases several illustrative examples of typical cases from the dataset. The dataset is randomly divided into separate subsets for training and validation. The partitioning follows an 80:20 ratio, with 80\% of the data assigned to training and 20\% to validation. The objective is to perform automated segmentation of the 3D facial wound region and integrate it with the findings of Phuong et al. \cite{Phuong_2023} regarding defect face reconstruction to extract the wound-filling part specific to the analyzed face.

\subsection{Training strategy}
\label{strategy}
\begin{algorithm}
\caption{\textbf{:} Training strategy (segmentation)}
\label{Training_strategy}
$\begin{array}{ll}
\hspace*{0.2cm}\textbf{Input} &\textbf{: } \mathcal{D} \gets \text{Datasets including training and validation}\\
 & \textbf{: } \mathcal{M} \gets \text{Architecture of the segmentation model } \text{\cite{Zhao_2022}}\\
 \hspace*{0.2cm} \textbf{Output} & \textbf{: } \mathcal{O} \gets \text{The wound segmentation model is most effective}
\end{array}$
\begin{algorithmic}
\State $ \mathcal{L}  \gets \text{[Focal\_Loss, Dice\_Loss, Cross\_Entropy\_Segmentation\_Loss, Weighted\_Cross\_Entropy\_Loss]}$
\State $N \gets$ length($\mathcal{L}$)
\State $\mathcal{P} \gets$ List[ ] \Comment{The list contains the best mIOU of the training models for each corresponding loss function}
\For{$k \gets 1$ to $N$} 
	\State{$best\_mIoU \gets 0$ }
	\For{$epoch \gets 1$ to $50$} 
		\State{$loss\_training \gets \mathcal{L}_{k}$}
	 	\State{$mIoU \gets \mathcal{M}_{loss\_training}(\mathcal{D})$}
		 \If{$mIoU > best\_mIoU$}
    		\State{$best\_mIoU \gets mIoU $}
		 \EndIf
	\EndFor
	\State{AddItem($\mathcal{P}, best\_mIoU$)}
\EndFor
\State \Return $\mathcal{O} \gets$ Select the model with the highest $best\_mIoU$ from $\mathcal{P}$.
\end{algorithmic}
\end{algorithm}
We use the model in study \cite{Zhao_2022} to segment the wound area on the patient's 3D face. This model demonstrates a remarkable capacity for accurately discriminating boundaries between regions harboring distinct classes. Our dataset comprises two distinct classes, namely facial abnormalities and normal regions. Due to the significantly smaller proportion of facial wounds compared to the normal area, an appropriate training strategy is necessary to address the data imbalance phenomenon effectively. To address this challenge, we utilize specific functions that effectively handle data imbalance within the semantic segmentation task. These functions include focal loss \cite{Lin_2017}, dice loss \cite{Sudre_2017}, cross-entropy loss and weighted cross-entropy loss \cite{Jadon_2020}.

\textit{1) Focal loss} is defined as:
\begin{equation}
\mathcal{L}_{\text{focal\_loss}}= -\alpha_t(1 - p_t)^\gamma \log(p_t),
\end{equation}
where $p_t$ represents the predicted probability of the true class; $\alpha_t$ is the balancing factor that assigns different weights to different classes; $\gamma$ is the focusing parameter that modulates the rate at which easy and hard examples are emphasized. Focal loss effectively reduces the loss contribution from well-classified examples and focuses on samples that are difficult to classify correctly. This helps handle class imbalance and improves the model's performance on minority classes.

\textit{2) Dice loss} also known as the Sorensen-Dice coefficient is defined as:
\begin{equation}
\mathcal{L}_{\text{dice\_loss}}=1-\frac{2 \sum_{i=1}^N p_i y_i + \epsilon}{\sum_{i=1}^N p_i^2+\sum_{i=1}^N y_i^2 + \epsilon},
\end{equation}
where $p$ represents the predicted probability or output of the model; $y$ is the ground truth or target labels; $N$ is the number of elements in the predicted and ground truth vectors; $\epsilon$ is a small constant added to the denominator to avoid division by zero.

\textit{3) Cross-entropy segmentation loss} is defined as:
\begin{equation}
\mathcal{L}_{\text{cross\_entropy\_segmentation\_loss}} =-\sum_{i=1}^M \sum_{c=1}^C y_{i c} \log \left(p_{i c}\right),
\end{equation}
where $y_{ic}$ denotes the ground truth label for the $i$-th sample and $c$-th class;  $p_{ic}$ represents the predicted probability for the $i$-th sample and $c$-th class; $M$ is the total number of samples; $C$ is the number of classes.

\textit{4) Weighted Cross-Entropy loss} is as follows:
\begin{equation}
\mathcal{L}_{\text{weighted\_cross\_entropy\_loss}}= -\frac{1}{N} \sum_{i=1}^N w r_i \log \left(p_i\right)+\left(1-r_i\right) \log \left(1-p_i\right),
\end{equation}
where $w = \dfrac{N-\sum_n p_n}{\sum_n p_n}$ represents the weight assigned to each point based on its class. 

The model undergoes training through iterative experimental minimization of the loss functions in order to select the most efficacious model. The training process was conducted using a single NVIDIA Quadro RTX 6000 GPU over the course of 50 epochs. The Adam optimizer is employed in conjunction with a mini-batch size of 4. The initial learning rate was set at $1e-3$, and it underwent a decay of 0.5 every 20 epochs. The experimental process for identifying the most efficient model for wound segmentation is meticulously elucidated in Algorithm \ref{Training_strategy} within the study.

\section{Results and discussion}
\begin{figure}[!h]
\centering
\includegraphics[scale=0.48]{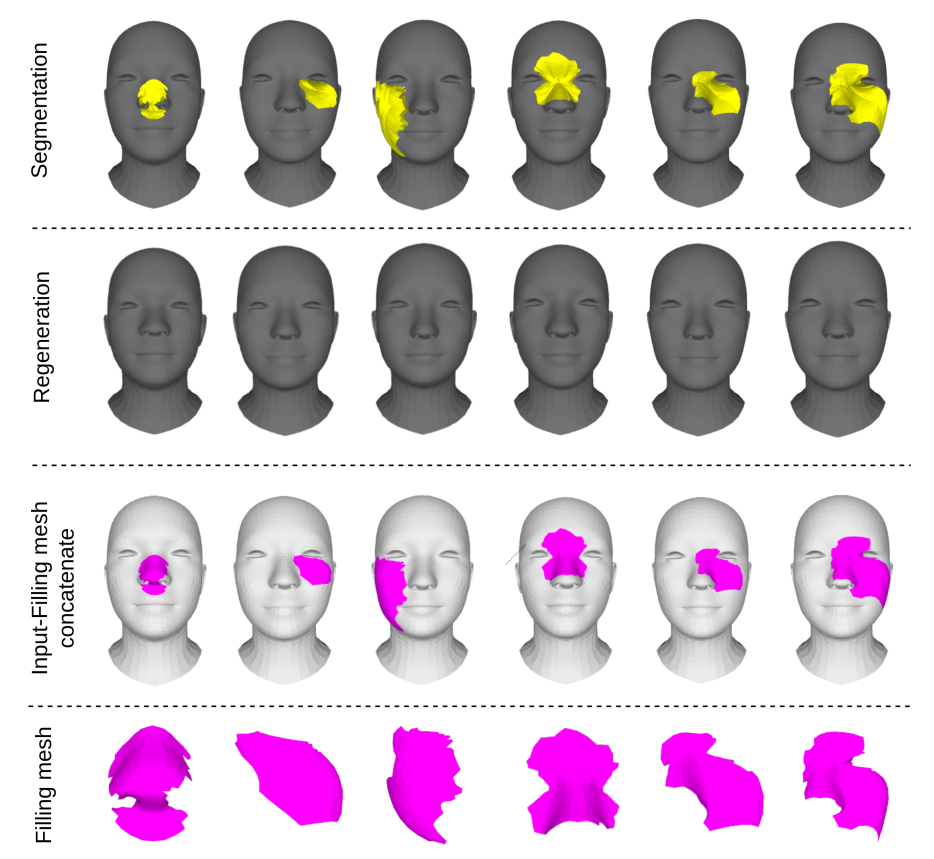}
\caption{Illustrate improvement of fill extraction results with 3D printing}
\label{results}
\end{figure}

The model was trained on the dataset using four iterations of experiments, wherein different loss functions were employed. The outcomes of these experiments are presented in Table \ref{result_loss_function}. The utilized loss functions demonstrate excellent performance in the training phase, yielding highly satisfactory outcomes on large-scale unbalanced datasets. Specifically, we observe that the model integrated with cross-entropy segmentation loss exhibits rapid convergence, requiring only 16 epochs to achieve highly favorable outcomes. As outlined in Section \ref{strategy}, the model exhibiting the most favorable outcomes, as determined by the cross-entropy segmentation loss function, was selected for the segmentation task. This particular model achieved an impressive mIoU score of 0.9999986. Some illustrations for the segmentation result on a 3D face are shown in Fig. \ref{results}. The results demonstrate the effectiveness of the two-stream graph convolutional network in accurately segmenting complex and minor wounds, indicating that the model successfully captures the geometric feature information from the 3D data.

\begin{table}[!h]
\renewcommand{\arraystretch}{1.5}
\centering
\caption{Results of training the model with the corresponding loss functions}
\label{result_loss_function}
\begin{tabular}{|c|c|c|}
\hline
Loss Function	& $mIoU$ & Number of epochs\\
\hline
Focal\_Loss	& 0.9998651 & 50 \\
\hline
Dice\_Loss	& 0.9969331 & 50 \\
\hline
Cross\_Entropy\_Segmentation\_Loss	& 0.9999986 & 16 \\
\hline
Weighted\_Cross\_Entropy\_Loss	& 0.9999863 & 50 \\
\hline
\end{tabular}
\end{table}

\begin{algorithm}
\caption{\textbf{:} Filling extraction (our proposal)}
\label{alg_filling_extraction}
$\begin{array}{ll}
\hspace*{0.2cm}\textbf{Input} &\textbf{: } \mathcal{M}_{in}(V,F)  \gets \text{ Mesh of injured patient's face}\\
 &\textbf{: } \mathcal{G} \gets \text{ Wound 3D facial regeneration model trained from research \cite{Phuong_2023}}\\
&\textbf{: }  \mathcal{S} \gets \text{Wound segmentation model according to the face of mesh}\\
\hspace*{0.2cm} \textbf{Output} & \textbf{: } \mathcal{M}_{extracted} \gets \text{Mesh of the wound filling part}
\end{array}$
\begin{algorithmic}
\State $\mathcal{M}_{\mathcal{G}}(V', F') \gets \mathcal{G}(\mathcal{M}_{in}(V,F))$ \Comment{Mesh of the regenerate face}
\State $\mathcal{F} \gets \mathcal{S}(\mathcal{M}_{in}(V,F))$ \Comment{Extract the injured labels in the mesh (faces) from the results of the segmentation model}
\State $\mathcal{V} \gets v(x,y,z) \in \mathcal{F}$ \Comment{Get vertices in $\mathcal{F}$}
\State $\mathcal{M}_{seg} \gets \text{ create mesh with vertices } \mathcal{V}$ and faces $\mathcal{F}$
\State $\mathcal{M}_{surface} \gets \mathcal{M}_{\mathcal{G}}(V', \mathcal{F})$ \Comment{create mesh with vertices $V'$ and faces $\mathcal{F}$}
\State $\mathcal{M}_{merged} \gets \mathrm{concatenate}(\mathcal{M}_{seg}, \mathcal{M}_{surface})$ \Comment{merges the two meshes into a single mesh, ensuring the faces and vertices are combined properly}
\State $\mathcal{M}_{extracted} \gets \mathrm{watertightness}(\mathcal{M}_{merged})$ \Comment{Clean up the mesh to ensure watertightness}
\State \Return  $\mathcal{M}_{extracted}$
\end{algorithmic}
\end{algorithm}

\begin{figure}[!h]
\centering
\includegraphics[scale=0.15]{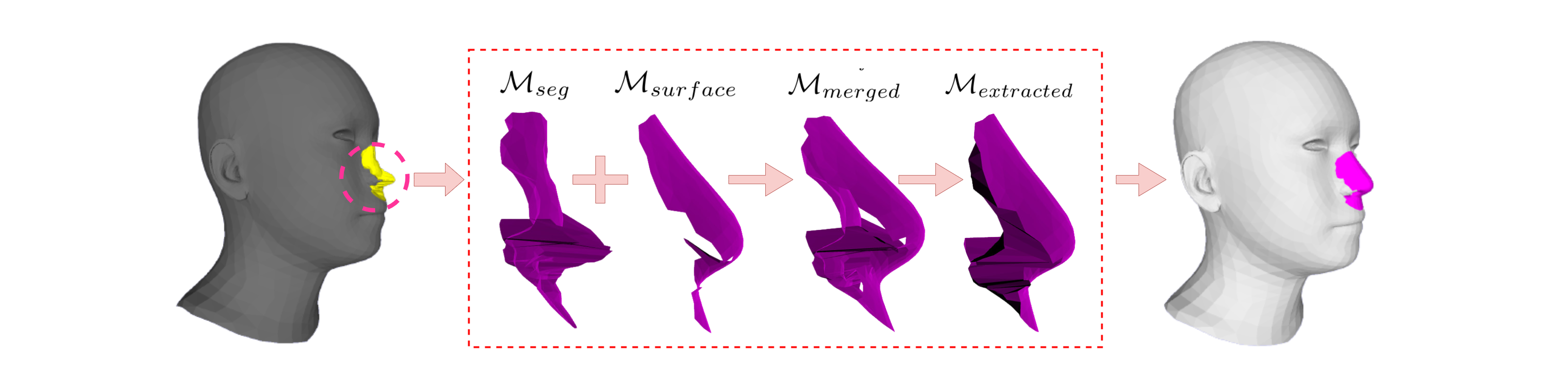}
\caption{Illustrate the wound-filling extraction algorithm}
\label{extract_filling}
\end{figure}

From the above segmentation result, our primary objective is to conduct a comparative analysis between our proposed wound fill extraction method and a method with similar objectives as discussed in the studies by Phuong et al. \cite{Phuong_2023, Phuong_2023_self}. A notable characteristic of the Cir3D-FaIR dataset is that all meshes possess a consistent vertex order. This enables us to streamline the extraction process of the wound filler. Utilizing the test dataset, we employ the model trained in the study by Phuong et al. \cite{Phuong_2023} for the reconstruction of the 3D face. Subsequently, we apply our proposed method to extract the wound fill from the reconstructed 3D face. Let $v(x,y,z) \in \mathcal{M}(V, F)$ is the vertices of the mesh containing the wound. In which $V$ and $F$ are the set of vertices and faces of the mesh, respectively. As previously stated, we introduce a methodology for the extraction of wound filling, which is comprehensively elucidated in Algorithm \ref{alg_filling_extraction} and Fig. \ref{extract_filling}.

\begin{algorithm}[!h]
\caption{\textbf{:} Compare the performance of the filling process}
\label{compare_filling_extraction}
$\begin{array}{ll}
\hspace*{0.2cm}\textbf{Input} &\textbf{: } \mathcal{D} = \{\mathcal{M}_{i}(V,F)\}^{N}_{i=1} \gets \text{The test dataset consists of N meshes}\\
\hspace*{0.2cm} \textbf{Output} & \textbf{: } \text{Performance of the outlier method and our proposed method}
\end{array}$
\begin{algorithmic}
\State $\mathcal{A}_{our\_proposal} \gets$ List[ ] \Comment{The list contains the accuracy of each mesh in our proposal}
\State $\mathcal{A}_{old} \gets$ List[ ]  \Comment{The list contains the accuracy of each mesh in the old proposal}
\For {$k \gets 1$ to $N$} 
	\State $\mathcal{M}_{extracted} \gets$ Wound filling part of $\mathcal{M}_{k}(V,F)$ from Algorithm \ref{alg_filling_extraction}
	\State $\mathcal{V}_{extracted} \gets$ Retrieve the index of the corresponding vertices of the faces in $\mathcal{M}_{extracted}$
	\State $\mathcal{V}_{GT} \gets$ Index of vertices with label as wound from $\mathcal{M}_{k}(V,F)$
	\State $\mathcal{V}_{old} \gets$ Index of vertices with wound is extracted from \textit{filling extraction algorithm} of research \cite{Phuong_2023}
	\State AddItem($\mathcal{A}_{our\_proposal}$, Accuracy($\mathcal{V}_{GT}, \mathcal{V}_{extracted}$))
	\State AddItem($\mathcal{A}_{old}$, Accuracy($\mathcal{V}_{GT}, \mathcal{V}_{old}$))
\EndFor
\State \Return average($\mathcal{A}_{old}$) and average($\mathcal{A}_{our\_proposal}$)
\end{algorithmic}
\end{algorithm}

\begin{figure}[!h]
\centering
\includegraphics[scale=0.13]{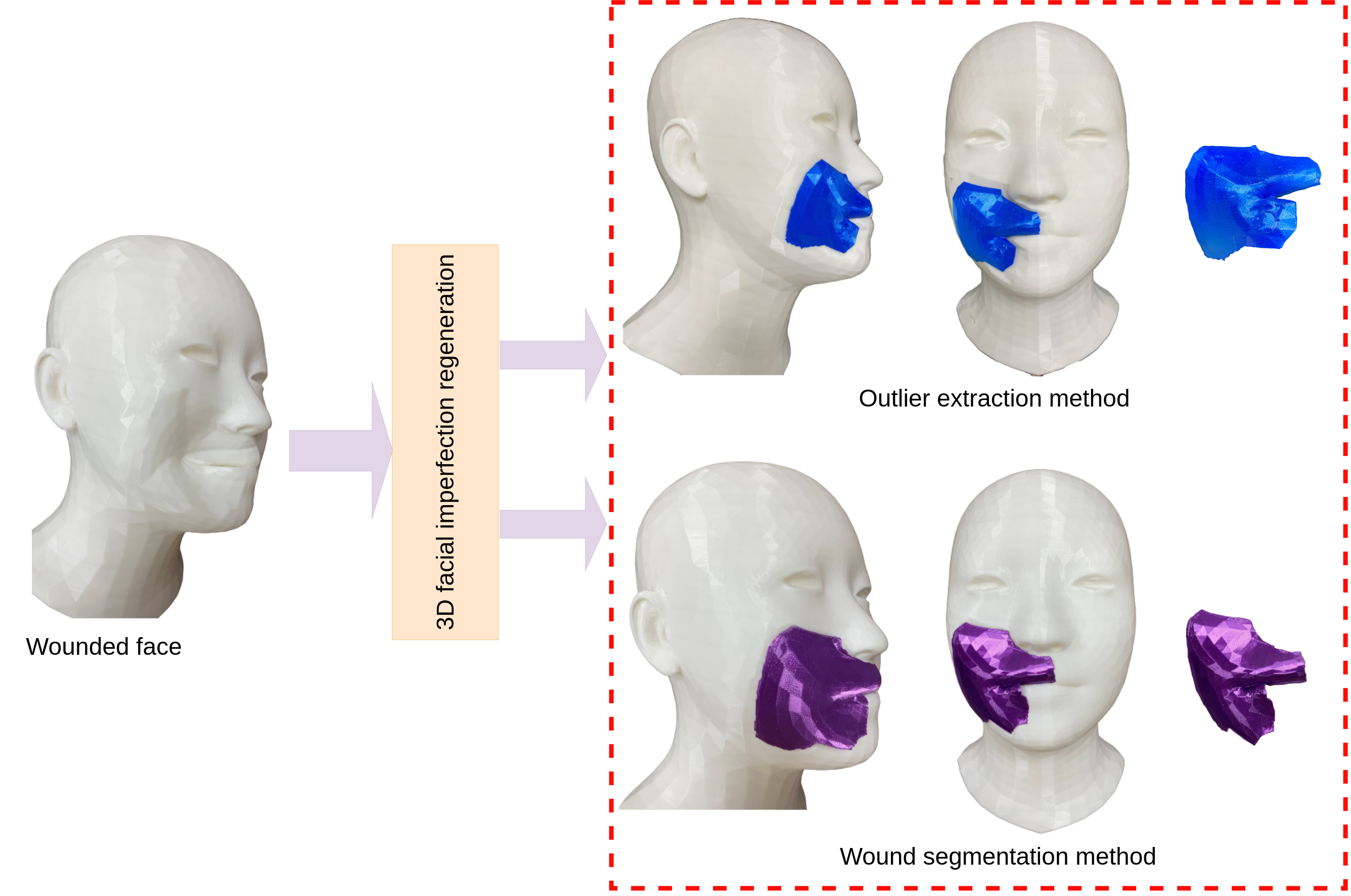}
\caption{Illustrate improvement of fill extraction results with 3D printing}
\label{compare_fix}
\end{figure}

For the purpose of notational convenience, we designate the filling extraction method presented in the study by Phuong et al. \cite{Phuong_2023} as the "old proposal". We conduct a performance evaluation of both our proposed method and the old proposal method on a dataset consisting of 8090 meshes, which corresponds to 20\% of the total dataset. A comprehensive description of the process for comparing the two methods is provided in Algorithm \ref{compare_filling_extraction}. The results show that our proposal has an average accuracy of \textbf{0.9999986\%}, while the method in the old proposal is \textbf{0.9715684\%}. The accuracy of the fill extraction method has been improved, which is very practical in the medical reconstruction problem. After that, the study randomly extracted the method outputs from the test set, depicted in Fig. \ref{results}. We have used 3D printing technology to illustrate the results of the actual model, which is significantly improved compared to the old method, as shown in Fig. \ref{compare_fix}. Our research only stops at proposing an efficient wound-filling extraction method with high accuracy. Therefore, further research can focus on developing personalized surgical planning tools based on reconstructed 3D models.

\section{Conclusions}
This study explored the benefits of using a two-stream graph convolutional network to segment 3D facial trauma defects automatically. Furthermore, we have proposed an improved method to extract the wound filling for the face. The results show the most prominent features as follows
\begin{itemize}
\item[-] An auto-segmentation model was trained to ascertain the precise location and shape of 3D facial wounds. We have experimented with different loss functions to give the most effective model in case of data imbalance. The results show that the model works well for complex wounds on the Cir3D-FaIR face dataset with an accuracy of 0.9999986\%.
\item[-] Concurrently, we have proposed a methodology to enhance wound-filling extraction performance by leveraging both a segmentation model and a 3D face reconstruction model. By employing this approach, we achieve higher accuracy than previous studies on the same problem. Additionally, this method obviates the necessity of possessing a pre-injury 3D model of the patient's face. Instead, it enables the precise determination of the wound's position, shape, and complexity, facilitating the rapid extraction of the filling material.
\item[-] This research proposal aims to contribute to advancing facial reconstruction techniques using AI and 3D bioprinting technology to print skin tissue implants. Printing skin tissue for transplants has the potential to revolutionize facial reconstruction procedures by providing personalized, functional, and readily available solutions. By harnessing the power of 3D bioprinting technology, facial defects can be effectively addressed, enhancing both cosmetic and functional patient outcomes. 
\item[-] From this research direction, our proposed approach offers a promising avenue for advancing surgical support systems and enhancing patient outcomes by addressing the challenges associated with facial defect reconstruction. Combining machine learning, 3D imaging, and segmentation techniques provides a comprehensive solution that empowers surgeons with precise information and facilitates personalized interventions in treating facial wounds.
\end{itemize}

\section*{Acknowledgments}
We would like to thank Vietnam Institute for
Advanced Study in Mathematics (VIASM) for hospitality during our visit in 2023, when we started to work on this paper.

\bibliographystyle{unsrt}  
\bibliography{3D_Seg} 

\end{document}